\documentclass[10pt,twocolumn,letterpaper]{article}

\usepackage{cvpr}              % To produce the CAMERA-READY version

\usepackage{float}
\usepackage{adjustbox}
\usepackage{booktabs}
\usepackage{multirow}
\usepackage{array}
\usepackage{graphicx}
\usepackage{subcaption}
\usepackage{xcolor}   % For coloring text
\usepackage[table]{xcolor}    % For better caption control
\usepackage{booktabs}     
\usepackage{siunitx}       % For aligning numbers in columns
\usepackage{threeparttable} % To add table notes for abbreviations
\usepackage{caption}       % For better caption control

\definecolor{cvprblue}{rgb}{0.21,0.49,0.74}
\usepackage[pagebackref,breaklinks,colorlinks,allcolors=cvprblue]{hyperref}

\def\paperID{41918} % *** Enter the Paper ID here
\def\confName{CVPR}
\def\confYear{2026}

\title{\centering PP-OCRv5: A Specialized 5M-Parameter Model Rivaling Billion-Parameter Vision-Language Models on OCR Tasks}

\author{
  Cheng Cui\thanks{The authors contributed equally.} \quad
  Yubo Zhang\footnotemark[1] \quad
  Ting Sun \quad
  Xueqing Wang \quad
  Hongen Liu \\[2pt]
  Manhui Lin \quad
  Yue Zhang \quad
  Tingquan Gao \quad
  Changda Zhou \quad
  Jiaxuan Liu \\[2pt]
  Zelun Zhang \quad
  Jing Zhang \quad
  Jun Zhang \quad
  Yi Liu\thanks{Corresponding author (liuyi22@baidu.com).} \\[3mm]
  PaddlePaddle Team, Baidu Inc.
}

\begin{document}
\maketitle
\begin{abstract}
The advent of “OCR 2.0” and large-scale vision-language models (VLMs) has set new benchmarks in text recognition. However, these unified architectures often come with significant computational demands, challenges in precise text localization within complex layouts, and a propensity for textual hallucinations. Revisiting the prevailing notion that model scale is the sole path to high accuracy, this paper introduces PP-OCRv5, a meticulously optimized, lightweight OCR system with merely 5 million parameters. We demonstrate that PP-OCRv5 achieves performance competitive with many billion-parameter VLMs on standard OCR benchmarks, while offering superior localization precision and reduced hallucinations. The cornerstone of our success lies not in architectural expansion but in a data-centric investigation. We systematically dissect the role of training data by quantifying three critical dimensions: data difficulty, data accuracy, and data diversity. Our extensive experiments reveal that with a sufficient volume of high-quality, accurately labeled, and diverse data, the performance ceiling for traditional, efficient two-stage OCR pipelines is far higher than commonly assumed. This work provides compelling evidence for the viability of lightweight, specialized models in the large-model era and offers practical insights into data curation for OCR. The source code and models are publicly available at \url{https://github.com/PaddlePaddle/PaddleOCR}.

% \textbf{Our code has been released and is open-source.}
\end{abstract}    
\section{Introduction}
\label{sec:intro}

The field of computer vision is increasingly dominated by large-scale models, particularly VLMs, which have demonstrated impressive generalist capabilities across a wide range of tasks~\cite{li2022blip, openai2023gpt4v, team2023gemini, bai2025qwen2}. This trend of scaling model size has naturally extended to OCR, with modern ``OCR 2.0'' and VLM-based approaches promising an end-to-end solution for extracting text from complex, real-world images. While the performance of these large models is academically remarkable, the prevailing focus on scale often encounters significant practical limitations when deployed in real-world OCR scenarios, which demand not only high recognition accuracy but also precision, reliability, and efficiency.

\begin{figure}[!t]
\centering
\includegraphics[width=\linewidth]{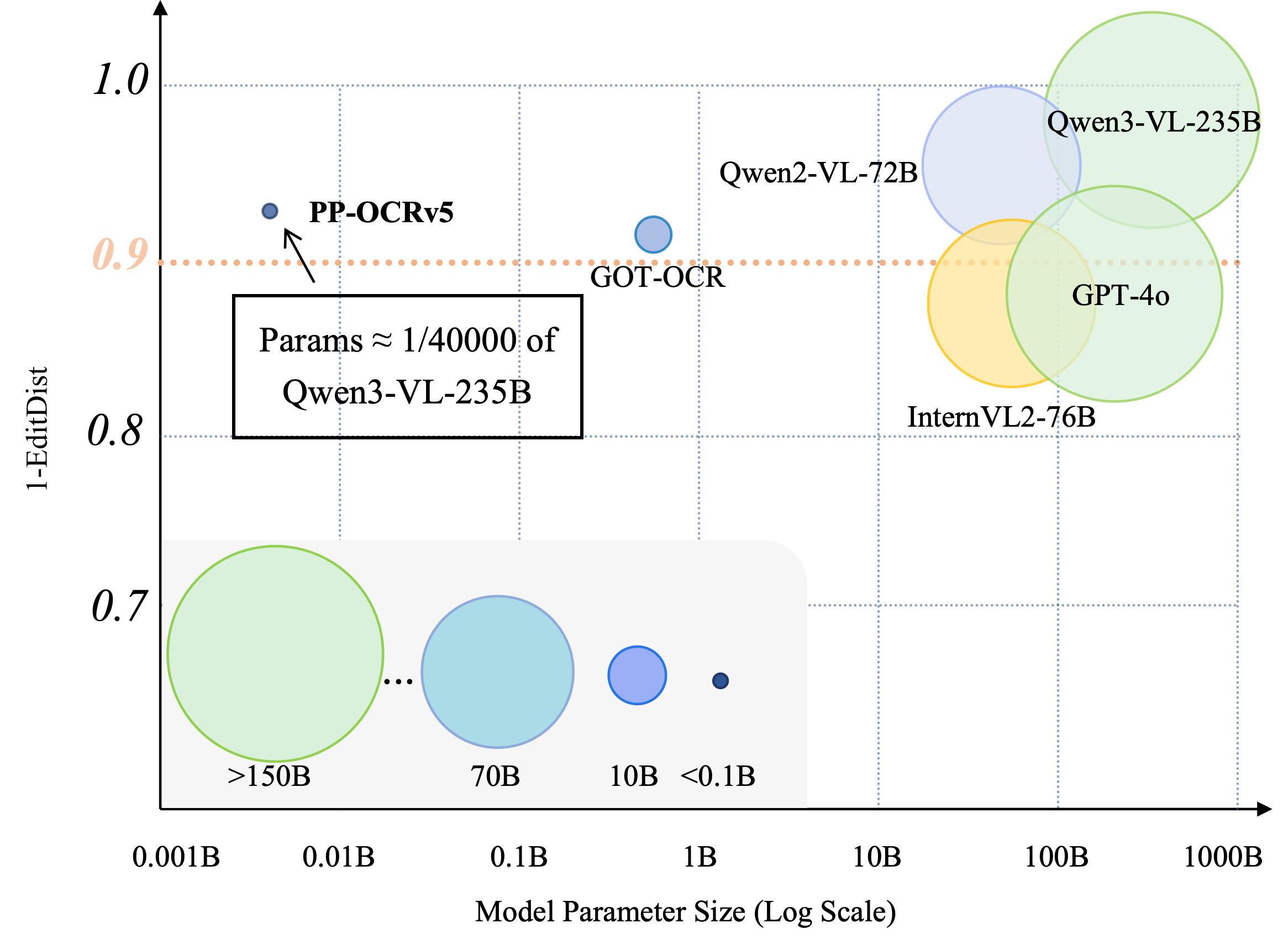} % 
\caption{
    PP-OCRv5 vs. Vision-Language Models: 1-EditDist Accuracy and Parameter Size Comparison. 
}
\label{fig:dataset}
\end{figure}

A primary challenge arises from the ``generalist's dilemma.'' Unified VLM architectures, although designed for diverse tasks, often lack the specialized capabilities needed for high-precision OCR. This manifests in three critical limitations:
(1) Imprecise Localization: These models often fail to produce the tight, accurate bounding boxes essential for document analysis, instead providing coarse region-of-interest indicators.
(2) Hallucination: In complex layouts, models can confidently ``hallucinate,'' generating plausible yet incorrect text. This presents a critical vulnerability for data-sensitive applications.
(3) Computational Inefficiency: Massive parameter counts render these models impractical for deployment in resource-constrained environments or for services demanding high-throughput, low-latency processing.
Collectively, these issues create a significant gap between large-scale model capabilities and industrial demands for fast, accurate, and reliable OCR.

These challenges inherent in large-scale models naturally steer the focus back towards lightweight, specialized OCR systems. However, the development of conventional small OCR models has also faced its own set of challenges, often hitting a performance plateau. A significant body of research has concentrated on model-centric innovations, introducing more sophisticated architectures for text detection and recognition~\cite{shi2016end, fang2021read}. While these architectural advancements have pushed benchmark performance, they often yield diminishing returns and are fundamentally constrained by the quality and scale of the training data. On the data front, while the use of synthetic data~\cite{gupta2016synthetic} and augmentation is standard practice, these methods are frequently applied in an ad-hoc manner, lacking a systematic framework to guide the curation and scaling of data. This results in models that, despite performing well on academic datasets, struggle to generalize to the vast “long-tail” of real-world scenarios, ultimately limiting their industrial applicability.

This exposes a critical research gap: the need for a methodology that can unlock the full potential of lightweight models, pushing their performance ceiling to rival that of large-scale counterparts. We therefore reconsider the prevailing model-centric narrative and ask a fundamental question: Can a meticulously optimized, lightweight model rival these giants in OCR scenarios? We posit that the performance ceiling of a model is determined not only by its parameter count but, more critically, by the difficulty, accuracy~\cite{song2022learning}, and diversity of its training data. This leads us back to a foundational principle: Data-Centric AI~\cite{sambasivan2021everyone, zha2025data, whang2023data}. We hypothesize that a specialized, efficient OCR model, when fueled by a massive, high-quality, and diverse dataset, can achieve performance comparable to, or even exceeding, its billion-parameter counterparts. To explore this, we move beyond treating data as a monolithic entity and instead propose a principled framework to analyze and scale it along three key dimensions: (1) Data Difficulty, identified by model confidence scores to filter out noise and overly simplistic samples; (2) Data Accuracy, quantifying the impact of label noise; and (3) Data Diversity, ensuring broad feature space coverage through systematic sampling.

Guided by this framework, we develop \textbf{PP-OCRv5}, a highly efficient two-stage (detection + recognition) OCR system with a total of only 5 million parameters\footnotemark. By applying our data-centric methodology to a large-scale dataset, we systematically enhance its capabilities. Our comprehensive experiments demonstrate that PP-OCRv5 achieves accuracy comparable to state-of-the-art billion-parameter VLMs on standard OCR benchmarks, while simultaneously providing superior localization precision, a drastic reduction in hallucinations, and a fraction of the computational cost.Our main contributions are summarized as follows:
\footnotetext{PP-OCRv5 is released in two variants: \textbf{PP-OCRv5\_mobile} (5M parameters), which serves as the primary subject of this paper, and \textbf{PP-OCRv5\_server}, a higher-capacity variant. Unless otherwise specified, all mentions of ``PP-OCRv5'' in this paper refer to PP-OCRv5\_mobile.}

\begin{itemize}
    \item We present PP-OCRv5, a practical, lightweight, and easily deployable OCR system that is highly efficient.
    \item We propose a novel and systematic guiding principle for data-centric scaling in OCR, which analyzes and optimizes large-scale training data from the perspectives of difficulty, accuracy, and diversity. This guiding principle is not only applicable to the development of small OCR models but may also provide valuable insights for the development of any expert models.
    \item We demonstrate that PP-OCRv5, optimized through our data-centric methodology, achieves competitive performance with billion-parameter VLMs while providing improved text localization precision, reduced computational cost, and lower susceptibility to hallucinations—making it a practical and efficient alternative for production OCR deployments.
\end{itemize}
\section{Related Work}
This section reviews key advancements in the field of OCR, focusing on two primary research directions: the conventional two-stage OCR pipeline, which has become an industry standard, and the emerging application of large-scale Vision-Language Models (VLMs) for text-related tasks.

\subsection{Specialized Two-Stage OCR Systems}

The two-stage pipeline, which decouples text detection and recognition, remains a cornerstone of high-performance OCR due to its proven accuracy and efficiency, making it the de facto standard in industrial applications. This approach typically combines a segmentation-based detector with a Transformer-based recognizer. For detection, methods like DBNet~\cite{liao2020real} and its variants (e.g., PANet~\cite{wang2019panet}) are favored for their robustness in handling arbitrarily shaped text via a differentiable binarization module. For recognition, the field has evolved from the seminal CRNN~\cite{shi2016end} to powerful Vision Transformer architectures like SVTR~\cite{du2022svtr}, which excel at capturing long-range dependencies. The PP-OCR series~\cite{du2020pp, li2022pp}, which serves as the foundation for our work, exemplifies this evolution by integrating these advanced components with aggressive efficiency-driven optimizations, such as lightweight backbones (e.g., PP-LCNet~\cite{cui2021pplcnetlightweightcpuconvolutional}) and knowledge distillation. However, these prior efforts primarily focused on architectural and algorithmic enhancements, treating the training data as a largely static component. Our PP-OCRv5 builds upon this strong lineage but represents a paradigm shift, moving the focus from architectural tweaks to a systematic, data-centric scaling methodology, which is the core contribution of this work.

\begin{figure*}[h]
\centering
    % 第一个子图
\includegraphics[width=\linewidth]{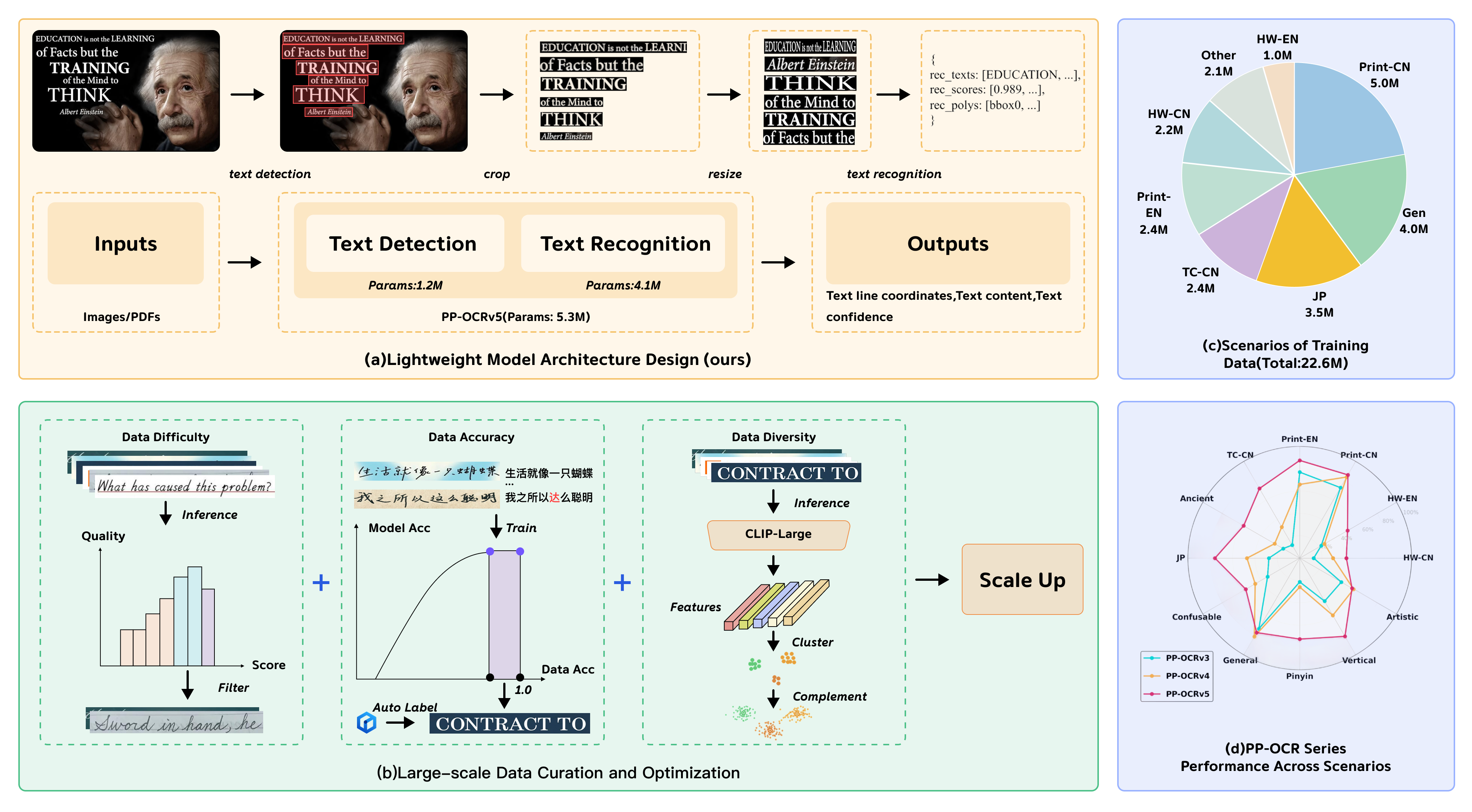} 

\caption{
    % \centering
    Overview of the PP-OCRv5 Framework: (a) Lightweight model architecture design; (b) Large-scale data construction and optimization strategies; (c) Distribution of training data across 22.6M samples; (d) Comparative performance of the PP-OCR series across multiple scenarios.
}
\label{fig:ocr_v5_pipeline}
\end{figure*}

\subsection{Vision-Language Models for OCR}

Recently, large-scale VLMs such as GPT-4V~\cite{yang2023dawn}, Gemini~\cite{team2023gemini}, and Qwen-VL~\cite{bai2025qwen2} have demonstrated unprecedented multimodal understanding capabilities. Pre-trained on vast amounts of image-text data, these models can perform OCR in a zero-shot or few-shot manner and comprehend the semantic and layout context of text within a document. This makes them exceptionally suited for high-level tasks like Document AI and Visual Question Answering (VQA), e.g., directly answering “What is the total amount on the invoice?” from an image.

However, when VLMs are applied to core OCR tasks that demand high precision, their inherent limitations become apparent. First, coarse localization: VLMs typically do not output accurate polygon-level bounding boxes, which is unacceptable for applications requiring structured data extraction. Second, text hallucination: In complex or low-quality images, VLMs may generate “hallucinated” text that does not exist in the image, compromising the reliability of the output. Finally, prohibitive computational cost: Their massive model size and immense inference requirements make them difficult to deploy at scale, especially on edge devices or in high-throughput cloud services. While some specialized models designed for document understanding (e.g., Donut~\cite{kim2022ocr}, Pix2Struct~\cite{lee2023pix2struct}) have partially mitigated these issues, they still lag significantly behind lightweight, specialized models in terms of deployment efficiency and cost, as they often remain as sizable, monolithic architectures that lack the granular optimization of a dedicated two-stage pipeline.

In summary, while VLMs have opened new frontiers, their challenges in accuracy, reliability, and efficiency for core OCR tasks underscore the continued importance and practical value of optimizing lightweight, efficient, and professional OCR solutions like PP-OCRv5.

\section{Method}

% \subsection{Dataset}

% \subsection{Overview}

We propose PP-OCRv5, a highly efficient two-stage OCR system that systematically enhances model capability through data-centric scaling principles. As illustrated in Figure~\ref{fig:ocr_v5_pipeline}, our framework is built on two pillars: (1) a lightweight model architecture inherited from PP-OCRv4, and (2) a novel, large-scale data curation and optimization process. Given the strong performance of the PP-OCRv4 text detection model, the primary focus of the PP-OCRv5 upgrade was on augmenting the training data for the text recognition model to address challenging cases. Consequently, the data construction methodologies detailed in Section~\ref{sec:data-construction} pertain specifically to the text recognition model.

\begin{figure*}[htbp]
\centering
\begin{subfigure}[b]{0.48\textwidth}
    \centering
    \includegraphics[width=\textwidth]{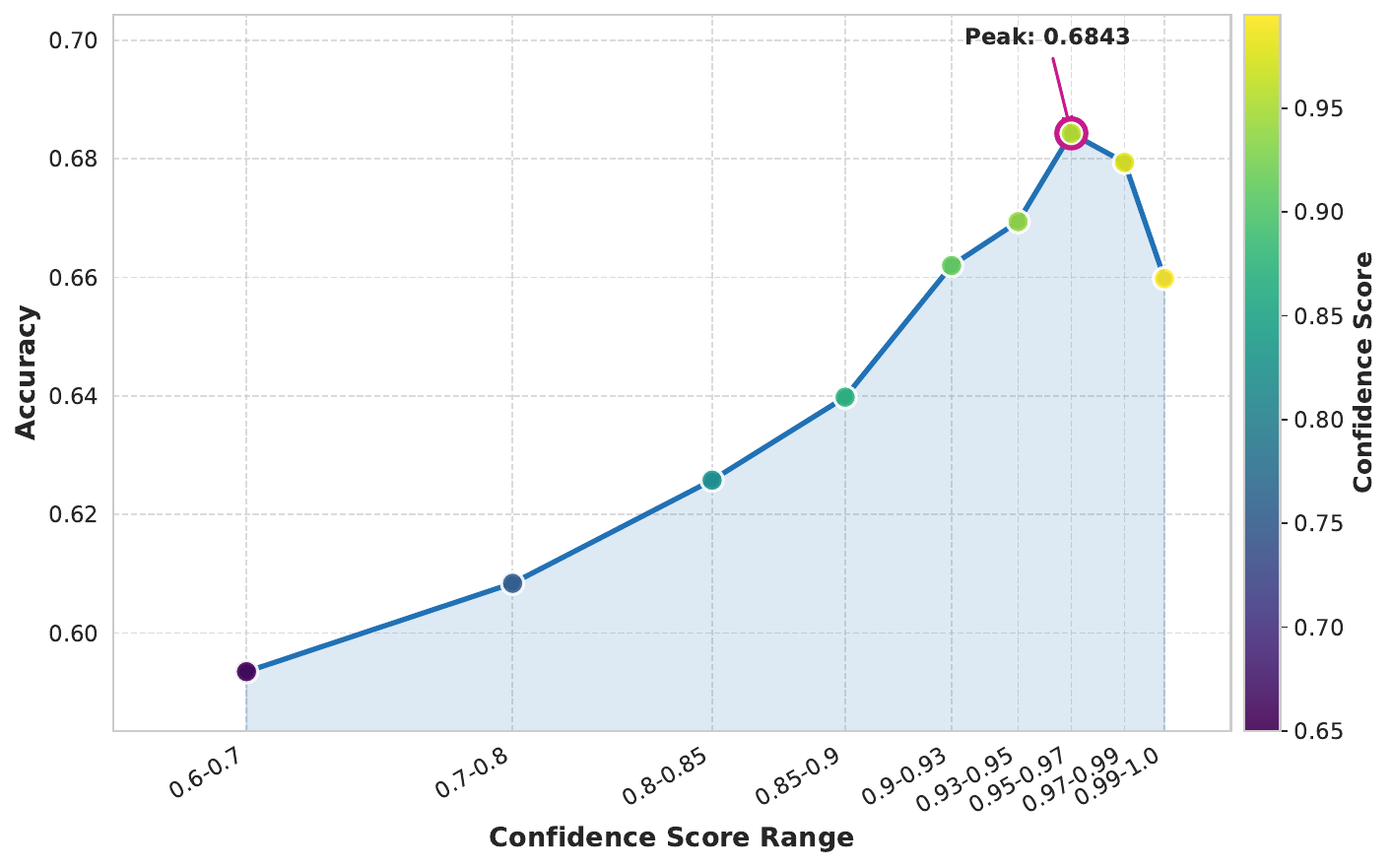}
    \caption{Impact of Data Difficulty on PP-OCRv5}
    \label{fig:dataset_1}
\end{subfigure}
\hfill
\begin{subfigure}[b]{0.48\textwidth}
    \centering
    \includegraphics[width=\textwidth]{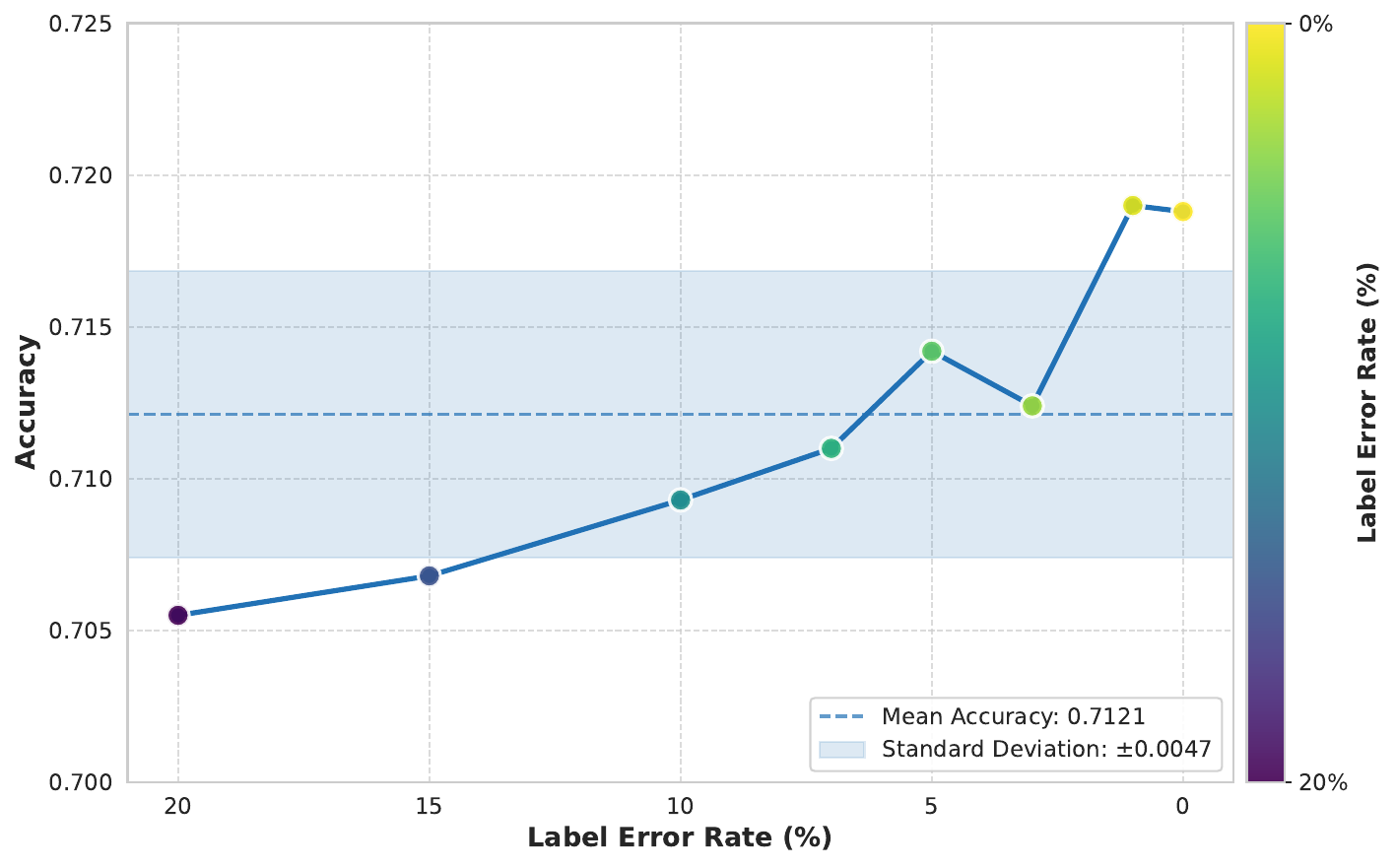}
    \caption{Impact of Data Accuracy on PP-OCRv5}
    \label{fig:dataset_2}
\end{subfigure}

\vspace{0.5cm}

\begin{subfigure}[b]{0.48\textwidth}
    \centering
    \includegraphics[width=\textwidth]{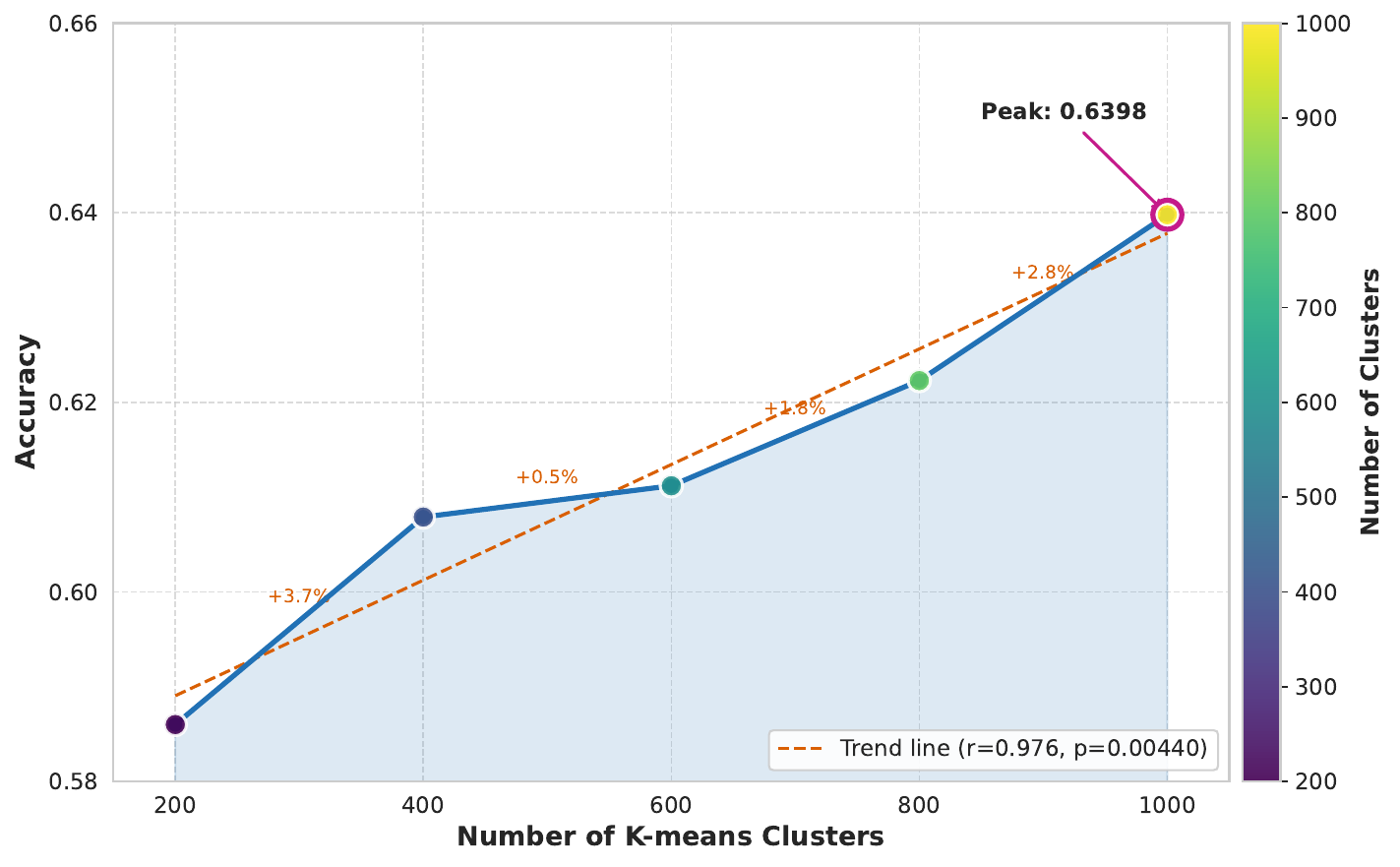}
    \caption{Impact of Data Diversity on PP-OCRv5}
    \label{fig:dataset_3}
\end{subfigure}
\hfill
\begin{subfigure}[b]{0.48\textwidth}
    \centering
    \includegraphics[width=\textwidth]{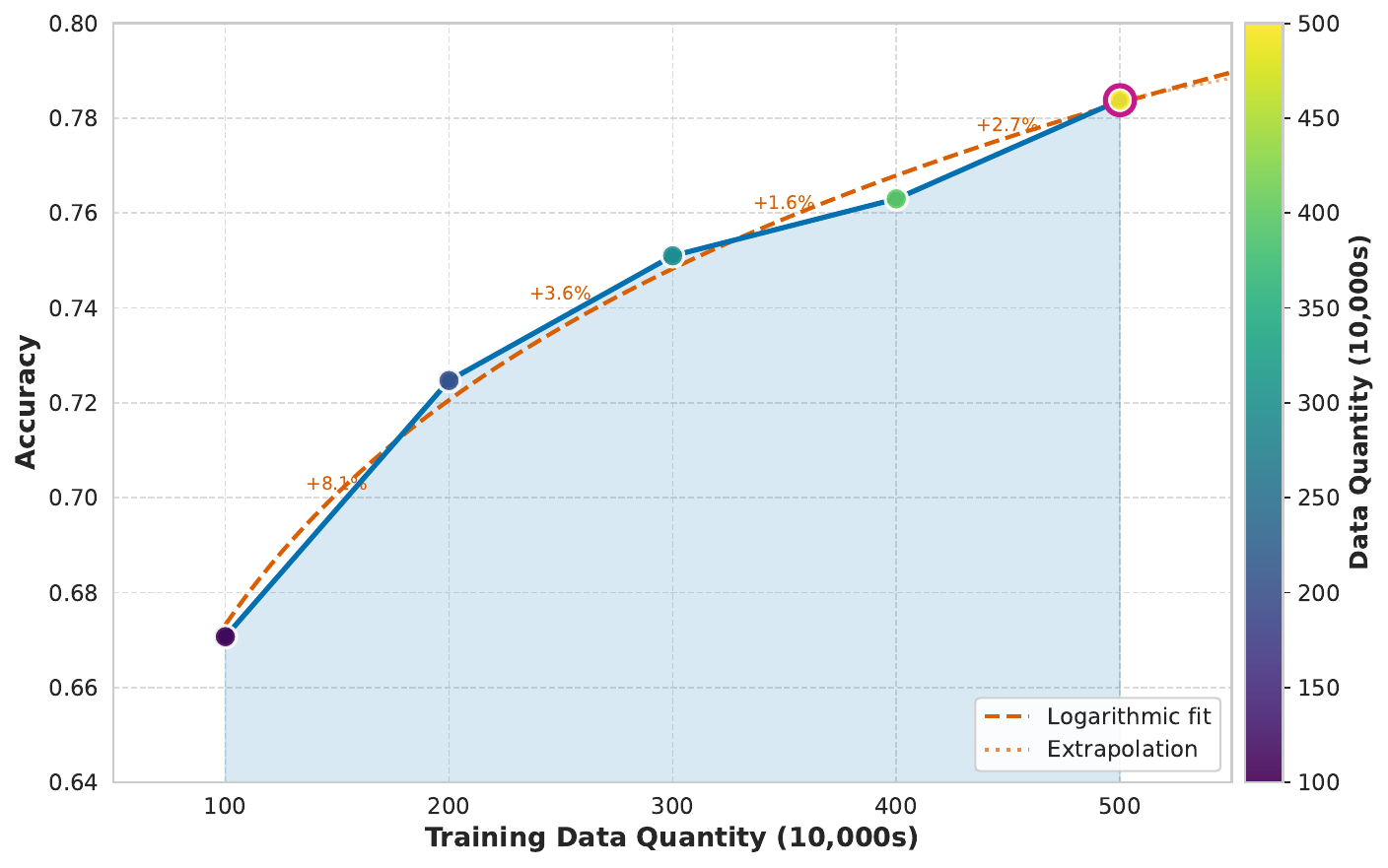}
    \caption{Impact of Data Quantity on PP-OCRv5}
    \label{fig:dataset_4}
\end{subfigure}

\caption{Comprehensive analysis of training data characteristics. Our key findings include: (a) A “sweet spot” of medium-difficulty data is most effective; (b) The model exhibits remarkable robustness to label noise; (c) Data diversity is paramount for performance; (d) Model accuracy scales nearly linearly with data quantity.}
\label{fig:dataset_combined}
\end{figure*}

\subsection{Lightweight Model Architecture Design}

Our architecture, summarized in Figure~\ref{fig:ocr_v5_pipeline}(a), follows the established and effective two-stage pipeline of PP-OCRv4. This design naturally incorporates a strong prior: textual information is predominantly organized in lines. The pipeline comprises a dedicated text detection model for locating text-line regions, and a text recognition model for decoding the content of the cropped lines. This division of labor simplifies the task for each model, allowing both to remain highly lightweight and efficient without sacrificing performance.

\subsubsection{Text Detection.} Our text detection module is built upon the DB algorithm~\cite{liao2020real} for robust text localization. It employs the PP-LCNetV3 backbone for efficient feature extraction. The network leverages a large kernel PAN~\cite{wang2019panet} structure as its neck, which is designed to capture features with a large receptive field, and a residual squeeze-and-excitation FPN~\cite{lin2017feature} that incorporates residual attention modules for multi-scale feature fusion. 

\subsubsection{Text Recognition.} The recognition model adopts SVTR\_LCNet~\cite{du2022svtr}, a lightweight hybrid architecture that combines the strengths of SVTR framework with PP-LCNetV3 backbone. The model employs the GTC~\cite{hu2020gtc} (Guided Training of CTC~\cite{graves2006connectionist}) strategy, where an attention-based decoder guides the CTC training process, enabling more effective sequence modeling and alignment. This approach leverages the complementary advantages of attention mechanisms for global dependency modeling and CTC for efficient sequence recognition. The integrated design ensures superior recognition accuracy for various text patterns while optimizing for deployment efficiency in resource-constrained environments.

\subsection{Large-scale Data Curation and Optimization}
\label{sec:data-construction}

To systematically optimize the training data for the PP-OCRv5 recognition model, we decompose the data along two principal dimensions: data quality and data quantity. With respect to data quality, we further abstract it into three key attributes: \textit{Data Difficulty}, \textit{Data Accuracy}, and \textit{Data Diversity}. For each attribute, we designed targeted experiments to quantitatively validate its impact on model accuracy, providing strong empirical support for developing ultra-lightweight yet high-precision text recognition models. In the following, we elaborate on the definitions and construction methodologies for each attribute. The overall workflow is depicted in Figure~\ref{fig:ocr_v5_pipeline}(b).

\subsubsection{Data Difficulty}

To quantify data difficulty, we first train a bootstrap recognition model on an initial $4$M-sample dataset, then apply it to candidate text lines. Each sample receives a confidence score $c \in [0, 1]$, computed as the average character-level softmax probability across the predicted sequence. A high score (\eg $c > 0.97$) indicates a visually simple text line, while a low score (\eg $c < 0.80$) typically signals either a genuinely hard sample or a mislabeled one. Samples are then ranked by $c$ to yield a continuous difficulty spectrum. Our ablation reveals a clear unimodal relationship: training on samples from the $[0.95, 0.97]$ range---challenging enough to provide informative gradients yet reliably labeled---yields the best performance, which we term the \textit{``sweet spot.''} This finding directly guided our final data sampling strategy.

\subsubsection{Data Accuracy}

To evaluate our model’s robustness to the label errors commonly found in real-world data, we conducted a controlled experiment with synthetic noise. Starting from a perfectly clean dataset, we created several training sets by injecting noise at predefined rates (e.g. 5\%, 10\%, 15\%). The noise was introduced by randomly selecting a percentage of samples and corrupting their labels, replacing 1-3 characters with random characters from the model’s vocabulary. This setup isolates the impact of label accuracy by keeping the image data distribution constant, allowing us to directly measure the performance degradation caused solely by label noise. This result provides a reference for determining the acceptable level of data accuracy in training datasets.

\subsubsection{Data Diversity}

Data diversity reflects the breadth of the visual feature space covered by the training corpus. As PP-OCRv5 targets not only document scenarios but also diverse wild scenes, we employ the visual encoder of CLIP~\cite{radford2021learning} as our feature extractor. Unlike domain-specific encoders, CLIP's general visual representations capture the semantic and stylistic variety inherent in real-world scenes (\eg text on a product label vs.\ a highway sign vs.\ a handwritten note), ensuring broad visual distribution coverage beyond merely varying text content. The extracted features are clustered into $1{,}000$ groups via K-Means~\cite{arthur2006k}, where each cluster approximates a distinct visual pattern. To isolate the effect of diversity from quantity, we construct five training sets of identical size ($600\text{k}$ samples) by sampling uniformly from $200$, $400$, $600$, $800$, and all $1{,}000$ clusters, respectively---increasing feature-space coverage while holding volume constant. This design enables a controlled measurement of diversity's contribution to model generalization.

The methodologies for \textbf{Data Difficulty}, \textbf{Data Accuracy}, and \textbf{Data Diversity} described above constitute a systematic experimental framework for acquiring high-quality training data. Through rigorous comparative analysis, we distill generalizable insights into data quality requirements for small-parameter models. By applying these data-centric strategies specifically to the development of PP-OCRv5, we constructed a refined training dataset with the distribution shown in Figure~\ref{fig:ocr_v5_pipeline}(c) and scaled the approach to larger volumes, ultimately obtaining the high-quality data essential for effective model training. As shown in Figure~\ref{fig:ocr_v5_pipeline}(d), PP-OCRv5 achieves the best accuracy in the series, demonstrating the effectiveness of our data-centric approach compared to previous versions.

\section{Experiments}

In this section, we present a rigorous evaluation of our data-centric framework and the resulting PP-OCRv5 model. All models in our ablation studies are trained and evaluated under consistent settings to ensure a fair comparison. Specifically, all experiments are conducted on 16 NVIDIA V100 GPUs with a batch size of 128. The models are trained for 100 epochs using the cosine learning rate scheduler, starting with a base learning rate of 0.0005 and 5 epochs of linear warm-up.

We first conduct in-depth ablation studies to quantitatively validate the impact of each dimension in our proposed framework—data difficulty, data accuracy, and data diversity—along with an investigation into data scaling laws (Section~\ref{Ablation Studies on Data Curation Strategy}). The empirical insights from these experiments directly informed the construction of our final, large-scale training dataset (Section~\ref{The PP-OCRv5 Training Dataset}). Finally, we demonstrate the highly competitive performance of the final PP-OCRv5 model through extensive comparisons against its predecessor and a suite of leading, large-scale, billion-parameter Vision-Language Models (VLMs) (Section~\ref{Comparison with State-of-the-Art}), while also highlighting its capabilities alongside notable efficiency.

\subsection{Ablation Studies on Data Curation Strategy}
\label{Ablation Studies on Data Curation Strategy}

To validate the effectiveness of the key components in our data curation framework, we conducted numerous detailed experiments based on the methods described in section \ref{sec:data-construction}, and the corresponding experimental results will be presented below.

\subsubsection{The Impact of Data Difficulty}

To validate our confidence-based difficulty ranking, we assessed its impact on model performance. We partitioned a 600k data subset into nine bins according to their pre-computed confidence scores and trained a PP-OCRv5 model independently on each. The results, illustrated in Figure \ref{fig:dataset_1}, reveal a distinct unimodal relationship.

Performance is suboptimal on low-confidence samples ($<0.8$), likely due to a higher prevalence of label noise. Conversely, accuracy peaks at 0.6843 in the 0.95-0.97 confidence range, which we identify as the “sweet spot” containing challenging yet correctly labeled samples. Notably, performance degrades on the highest-confidence samples ($>0.97$), suggesting these trivial examples contribute minimally to model generalization. 

This finding leads to a crucial insight for data strategy: blindly increasing dataset size is not the optimal path to improved model performance. Instead, curating a dataset from a ‘sweet spot’ of confidence—avoiding both the noisiest low-confidence samples and the most trivial high-confidence ones—is key. This selective approach enables the training of a more robust and generalizable model on a smaller, more efficient dataset, thereby significantly reducing computational costs and training time without sacrificing capability. As shown in Table~\ref{tab:conf_dist}, the final dataset is deliberately concentrated in the $0.95$--$0.97$ confidence range ($48.5\%$), reflecting the empirically validated ``sweet spot,'' while retaining a modest proportion of harder samples ($<0.90$, $13.6\%$ combined) to maintain robustness.

\begin{table}[h]
  \centering
  \footnotesize
  \setlength{\tabcolsep}{2pt}
  \vspace{-2mm}
  \begin{tabular}{lcl}
    \toprule
    Score Range & Ratio & Description \\
    \midrule
    $<$ 0.80 & 3.2\% & Hard / Ambiguous \\
    0.80-0.90 & 10.4\% & Challenging \\
    0.90-0.95 & 25.1\% & High Quality \\
    \textbf{0.95-0.97} & \textbf{48.5\%} & \textbf{Sweet Spot (Optimal)} \\
    $>$ 0.97 & 12.8\% & Simple (Stability) \\
    \bottomrule
  \end{tabular}
  % \vspace{-3mm}
  \caption{Confidence Score Distribution of the 22.6M Training Set.}
  \label{tab:conf_dist}
  % \vspace{-6mm}
\end{table}

\begin{table}[!htpb]
\centering
\small
\begin{tabular}{l r}
\toprule
\multicolumn{1}{c}{\textbf{Category}} & \textbf{Size (10,000s)} \\
\midrule
\textbf{1. Printed \& General Scenes} & \\
\quad Printed Chinese & 5,000 \\
\quad Printed English & 2,390 \\
\quad General Scenes & 4,010 \\
\quad Single Characters & 540 \\
\quad Special Formats & 50 \\
\midrule
\textbf{2. Handwritten} & \\
\quad Handwritten Chinese & 2,200 \\
\quad Handwritten English & 1,000 \\
\midrule
\textbf{3. Cross-lingual \& Scripts} & \\
\quad JP (Japanese) & 3,510 \\
\quad Traditional Chinese & 2,390 \\
\quad Ancient Books & 960 \\
\midrule
\textbf{4. Challenging Variations} & \\
\quad Confusable Characters & 100 \\
\quad Pinyin & 70 \\
\quad Rare Characters & 280 \\
\quad Vertical Text & 30 \\
\quad Artistic Text & 1.5 \\
\quad Emoji & 20 \\
\midrule
\textbf{Total Dataset Size} & \textbf{22,551.5} \\
\bottomrule
\end{tabular}
\caption{Statistics of the PP-OCRv5 training dataset. The data is categorized into four major groups, with a total of approximately 22.6M samples.}
\label{tab:dataset-statistics}
\vspace{-6pt}
\end{table}

\begin{table*}[!htbp]
\centering
\footnotesize
\begin{tabular}{@{}l c *{12}{c} @{}}
\toprule
\multirow{3}{*}{\textbf{Model}} & \multirow{3}{*}{\textbf{Weighted Acc.}} & \multicolumn{12}{c}{\textbf{Scenario-specific Accuracy (\%)}} \\
\cmidrule(l){3-14}
 & & \multicolumn{2}{c}{\textbf{Handwritten}} & \multicolumn{2}{c}{\textbf{Printed}} & \multirow{2}{*}{\textbf{TC-CN}} & \multirow{2}{*}{\textbf{Ancient}} & \multirow{2}{*}{\textbf{JP}} & \multirow{2}{*}{\textbf{Confus.}} & \multirow{2}{*}{\textbf{Gen.}} & \multirow{2}{*}{\textbf{Pinyin}} & \multirow{2}{*}{\textbf{Vert.}} & \multirow{2}{*}{\textbf{Art.}} \\
\cmidrule(l){3-6}
 & & \textbf{CN} & \textbf{EN} & \textbf{CN} & \textbf{EN} & & & & & & & & \\
\midrule
PP-OCRv3 & 42.5 & 12.5 & 22.2 & 72.9 & 77.0 & 13.5 & 17.1 & 27.6 & 33.2 & 73.1 & 21.3 & 44.5 & 42.9 \\
PP-OCRv4 & \cellcolor{cyan!10}{53.0} & \cellcolor{cyan!10}{29.8} & \cellcolor{cyan!10}{25.5} & \cellcolor{cyan!10}{83.9} & \cellcolor{cyan!10}{66.0} & \cellcolor{cyan!10}{32.2} & \cellcolor{cyan!10}{25.9} & \cellcolor{cyan!10}{47.2} & \cellcolor{cyan!10}{46.0} & \cellcolor{red!10}81.1 & \cellcolor{cyan!10}{25.9} & \cellcolor{cyan!10}{59.2} & \cellcolor{cyan!10}{55.6} \\
\textbf{PP-OCRv5} & \cellcolor{red!10}\textbf{80.1} & \cellcolor{red!10}\textbf{41.7} & \cellcolor{red!10}\textbf{49.4} & \cellcolor{red!10}\textbf{86.1} & \cellcolor{red!10}\textbf{87.5} & \cellcolor{red!10}\textbf{72.0} & \cellcolor{red!10}\textbf{57.9} & \cellcolor{red!10}\textbf{75.8} & \cellcolor{red!10}\textbf{55.7} & \cellcolor{cyan!10}\textbf{77.0} & \cellcolor{red!10}\textbf{72.5} & \cellcolor{red!10}\textbf{80.9} & \cellcolor{red!10}\textbf{54.0} \\
\bottomrule
\end{tabular}
\caption{Text recognition accuracy (\%) of PP-OCR series models on our in-house benchmark, a comprehensive multi-scenario benchmark. The weighted accuracy (Weighted Acc.) is the primary overall metric. PP-OCRv5 demonstrates superior robustness across most scenarios. \colorbox{red!10}{Red} and \colorbox{cyan!10}{blue} cells indicate the best and second-best results per column, respectively. \textbf{Bold} values highlight PP-OCRv5 results.}

\label{tab:pp-ocrv5-recognition-results}

\vspace{4pt}
\footnotesize
\textbf{Abbreviations}: CN: Chinese, EN: English, TC: Traditional Chinese, JP: Japanese, Gen.: General Scenes, Confus.: Confusable, Vert.: Vertical, Art.: Artistic.
\end{table*}

\subsubsection{The Impact of Data Accuracy}

To quantify the model’s tolerance to label errors, we executed the synthetic noise experiment described in our methodology. The results, presented in Figure \ref{fig:dataset_2}, reveal the model’s exceptional resilience.

Specifically, as the label accuracy was systematically reduced from 100\% (a pristine dataset) to 80\% (a 20\% corruption rate), the model’s recognition accuracy only decreased from 0.7188 to 0.7055. This represents a performance drop of merely 1.33 percentage points, despite a substantial increase in label noise. This remarkable stability suggests that the model learns robust visual features from the image content itself, effectively mitigating the impact of incorrect supervisory signals. 

This characteristic leads to an important conclusion: when training such small models, we can tolerate some noise in the labels. This makes it possible for large vision-language models to automatically annotate massive datasets—although these models may occasionally produce some unusual errors during the annotation process, the error rate does not significantly affect the accuracy of the small model. As a result, the cost of data annotation for training small models can be greatly reduced.

\subsubsection{The Impact of Data Diversity}

We executed the data diversity experiment to isolate the impact of feature richness while keeping data volume constant. The results, depicted in Figure \ref{fig:dataset_3}, establish a clear and monotonic relationship between the number of source clusters and model accuracy. As the diversity of the 600k training set increased from 200 to 1000 clusters, the model’s accuracy rose substantially from 0.5860 to 0.6398, a gain of 5.38 percentage points.

This finding highlights a fundamental principle of efficient model training: feature diversity, not mere data volume, is the true engine of generalization. Our results demonstrate that a model’s performance ceiling is determined by the breadth of the feature space it has been exposed to. Simply accumulating more samples from familiar patterns leads to diminishing returns, as the model gains little new information. In contrast, strategically acquiring data to cover novel feature clusters forces the model to learn more robust and comprehensive representations. This establishes a clear directive for data strategy: prioritize diversity-driven acquisition over volume-centric accumulation.This principle guided the curation of our final 22.6M dataset.

\subsubsection{The Impact of Data Quantity}

Having established the foundational role of data diversity, we next investigated its synergistic relationship with data quantity. We trained the model on datasets of increasing scale, from 1 million to 5 million samples, ensuring each subset was drawn from our established diverse data pool. 

As illustrated in Figure \ref{fig:dataset_4}, quintupling the training data resulted in a massive 11.3-point leap in accuracy, from 0.6707 to 0.7838. This nearly linear improvement demonstrates that diversity unlocks the potential of scale; a model that has learned a robust representation from a diverse feature space possesses a strong base, allowing it to effectively consolidate and fine-tune its knowledge with more examples. This result does not contradict our previous finding but rather complements it, providing a clear rationale for a two-fold data strategy: first, establish broad feature-space coverage, and second, scale the data volume within that diverse foundation to maximize model performance.

\subsection{The PP-OCRv5 Training Dataset}
\label{The PP-OCRv5 Training Dataset}

Guided by the insights from our comprehensive data experiments, we constructed the final 22.6M training dataset, which serves as the cornerstone for PP-OCRv5’s performance. The composition of this dataset is not arbitrary but a direct implementation of our empirically validated data-centric principles.

Our investigation into data difficulty, accuracy, diversity, and scale provided a clear blueprint. To broaden PP-OCRv5’s application scope, we incorporated a multi-source composition covering diverse scenarios from general print to challenging artistic text. Crucially, our diversity experiment proved that feature-space richness within each category is paramount. Therefore, for each category, we sourced data with a wide variety of intrinsic features. The data scale experiment then justified expanding this high-quality, feature-rich collection to 22.6M samples.

As detailed in Table \ref{tab:dataset-statistics}, the dataset is strategically assembled, with data volume distributed according to task difficulty and practical needs. During training, dynamic sampling strategies (e.g., oversampling) are applied to balance learning across different categories, ensuring the model effectively learns from all types of data. This approach allows us to build a model that is both highly competent in core domains and robust across a wide array of long-tail scenarios.

\subsection{Comparison with State-of-the-Art}
\label{Comparison with State-of-the-Art}

To comprehensively evaluate the performance of our model, we conduct a series of rigorous experiments. We compare our model against its direct predecessors on our in-house evaluate dataset, assess its capabilities relative to leading trillion-parameter scale vision-language models, and finally, benchmark it against a wide array of specialized tools and large models on the public OmniDocBench dataset.

\begin{table*}[!h]
    \centering
    \resizebox{\textwidth}{!}{%
    \renewcommand{\arraystretch}{1.2}
    \begin{tabular}{l|l|c|c|ccc|ccc|cccc}
        \toprule
        \multirow{2}{*}{\textbf{Model Type}} & \multirow{2}{*}{\textbf{Model}} & \multirow{2}{*}{\textbf{Params}} & \multirow{2}{*}{\textbf{ALL\_avg}$\downarrow$} & \multicolumn{3}{c|}{\textbf{Language}} & \multicolumn{3}{c|}{\textbf{Text Background}} & \multicolumn{4}{c}{\textbf{Text Rotate}} \\
        \cmidrule(lr){5-7} \cmidrule(lr){8-10} \cmidrule(lr){11-14}
        & & & & \textit{EN}$\downarrow$ & \textit{ZH}$\downarrow$ & \textit{Mixed}$\downarrow$ & \textit{White}$\downarrow$ & \textit{Single}$\downarrow$ & \textit{Multi}$\downarrow$ & \textit{Normal}$\downarrow$ & \textit{Rotate90}$\downarrow$ & \textit{Horizontal}$\downarrow$ & \textit{Rotate270}$\downarrow$ \\
        \midrule
        \multirow{4}{*}{\textbf{\begin{tabular}{c}Vision Language\\Models\end{tabular}}} 
        & Qwen3-VL & 235B & \cellcolor{red!15}{0.026} & \cellcolor{red!15}{0.016} & \cellcolor{red!15}{0.026} & \cellcolor{red!15}{0.034} & \cellcolor{red!15}{0.023} & \cellcolor{red!15}{0.023} & \cellcolor{red!15}{0.030} & \cellcolor{red!15}{0.023} & 0.046 & \cellcolor{red!15}{0.029} & \cellcolor{cyan!15}{0.170} \\
        & Qwen2-VL & 72B & 0.173 & 0.072 & 0.274 & 0.286 & 0.234 & 0.155 & 0.148 & 0.223 & 0.273 & 0.067 & 0.721 \\
        & InternVL2 & 76B & 0.115 & 0.074 & 0.155 & 0.242 & 0.113 & 0.352 & 0.269 & 0.132 & 0.610 & 0.595 & 0.907 \\
        & GPT4o & - & 0.122 & \cellcolor{cyan!15}{0.020} & 0.224 & 0.125 & 0.167 & 0.140 & 0.220 & 0.168 & 0.115 & 0.132 & 0.718 \\
        \midrule
        \multirow{6}{*}{\textbf{\begin{tabular}{c}Pipeline Tools\\\&\\Expert Vision\\Models\end{tabular}}} 
        & Tesseract-OCR & - & 0.324 & 0.096 & 0.551 & 0.250 & 0.439 & 0.328 & 0.331 & 0.426 & 0.117 & 0.984 & 0.969 \\
        & EasyOCR & - & 0.329 & 0.260 & 0.398 & 0.445 & 0.366 & 0.287 & 0.388 & 0.360 & 0.970 & 0.926 & 0.997 \\
        & Surya & - & 0.090 & 0.057 & 0.123 & 0.164 & 0.093 & 0.186 & 0.235 & 0.104 & 0.634 & 0.255 & 0.767 \\
        & Mathpix & - & 0.137 &  0.033 & 0.240 & 0.261 & 0.185 & 0.121 & 0.166 & 0.180 & \cellcolor{cyan!15}{0.038} & 0.638 & 0.185 \\
        & GOT-OCR & 0.58B & 0.077 & 0.041 & 0.112 & 0.135 & 0.092 & \cellcolor{cyan!15}{0.052} & 0.155 & 0.091 & 0.562 & 0.097 & 0.966 \\
        & \textbf{PP-OCRv5} & 5M & \cellcolor{cyan!15}\textbf{0.067} & \textbf{0.058} & \cellcolor{cyan!15}\textbf{0.076} & \cellcolor{cyan!15}\textbf{0.103} & \cellcolor{cyan!15}\textbf{0.070} & \textbf{0.081} & \cellcolor{cyan!15}\textbf{0.095} & \cellcolor{cyan!15}\textbf{0.072} & \cellcolor{red!15}\textbf{0.012} & \cellcolor{cyan!15}\textbf{0.058} & \cellcolor{red!15}\textbf{0.139} \\
        \bottomrule
    \end{tabular}%
    }
   \caption{Component-level OCR text recognition evaluation on OmniDocBench. Results are reported by OmniDocBench~\cite{ouyang2024omnidocbenchbenchmarkingdiversepdf} except PP-OCRv5 and Qwen3-VL. \colorbox{red!15}{Red} and \colorbox{cyan!15}{blue} cells indicate the best and second-best results per column, respectively. \textbf{Bold} values highlight PP-OCRv5 results.}
   \label{tab:ocr_comprehensive_evaluation}
\end{table*}

\subsubsection{Performance on In-house Benchmark}

To quantify the advancements of our latest model, we conducted a rigorous comparison of the PP-OCR series, evaluating our new PP-OCRv5 against its immediate predecessors, PP-OCRv4 and PP-OCRv3 The evaluation was performed on our comprehensive in-house benchmark, which is specifically designed to cover a wide array of challenging real-world text recognition scenarios. This benchmark includes 12 distinct categories, ranging from handwritten and printed text in multiple languages to difficult cases like vertical text, artistic fonts, and confusable characters. The performance is measured by recognition accuracy (\%), with detailed results presented in Table \ref{tab:pp-ocrv5-recognition-results}.

In summary, PP-OCRv5 delivers a significant leap in recognition performance, raising the overall weighted average accuracy from 53.0\% (PP-OCRv4) to 80.1\%. The model achieves substantial breakthroughs in challenging scenarios such as handwriting, multilingual text, specialized formats, and complex layouts, with notable gains in handwritten English, Traditional Chinese, Japanese, Pinyin, and ancient book texts. Core capabilities in printed text recognition, especially for printed English, have also been markedly enhanced.

\subsubsection{Performance on OmniDocBench}

To objectively evaluate performance, we conducted a comprehensive comparison on the OmniDocBench~\cite{ouyang2024omnidocbenchbenchmarkingdiversepdf} public benchmark. Unlike OCRBench~\cite{liu2024ocrbench, fu2024ocrbench}, which evaluates OCR capabilities of LMMs on high-level tasks like VQA, OmniDocBench is more fitting for benchmarking PP-OCRv5 as a specialized text recognition engine, with its focus purely on recognition accuracy. Our evaluation includes PP-OCRv5, other leading specialized OCR engines, and the latest large-scale VLMs such as GPT-4o and Qwen2-VL-72B. We employ the Normalized Edit Distance as the core metric, where lower scores signify superior performance.

As shown in Table \ref{tab:ocr_comprehensive_evaluation}, PP-OCRv5 achieves state-of-the-art performance among specialized OCR models on the overall metric (ALL\_avg) with an average edit distance of 0.067. While very-large-scale multimodal models (VLMs) like Qwen3-VL achieve even lower scores, PP-OCRv5 demonstrates a favorable balance of high accuracy and practical efficiency for dedicated OCR tasks. The superiority of PP-OCRv5 is particularly evident in:

\begin{itemize}
\item \textbf{Multi-language Processing}: PP-OCRv5 maintains consistent performance across English (0.058), Chinese (0.076), and mixed-language (0.103) texts
\item \textbf{Challenging Layouts}: The model demonstrates exceptional robustness on rotated text (Rotate90: 0.012, Rotate270: 0.139), where VLMs and other OCR engines show significant degradation
\item \textbf{Complex Backgrounds}: PP-OCRv5 handles various background complexities effectively, outperforming competitors across all background types
\end{itemize}

While VLMs demonstrate promising capability on standard English text, they generally lack the high precision and robustness required for diverse and complex document OCR tasks. These results confirm that PP-OCRv5 serves as a highly effective and practical solution for real-world document recognition scenarios.

\section{Conclusion}

In this work, we demonstrate that a lightweight OCR model with only 5M parameters can rival 100B-parameter VLMs through a systematic, data-centric optimization strategy.

The cornerstone of our success is not merely amassing large quantities of data, but a meticulous approach to curating high-quality data. Through rigorous experimentation, we identified an optimal sample difficulty sweet-spot using prediction confidence, quantified the model's resilience to label noise, and proved that feature diversity, measured via clustering, is a critical driver of performance independent of data volume. This methodology enabled PP-OCRv5 to achieve performance on par with leading large-scale models, as evidenced by its strong results on OmniDocBench.

Our findings challenge the prevailing ``bigger is better'' paradigm, demonstrating that compact specialized models, when trained on data carefully curated across difficulty, accuracy, and diversity, can close a significant part of the performance gap against orders-of-magnitude larger counterparts—offering a clear and reproducible path toward OCR systems that are highly efficient, accurate, and readily deployable in real-world scenarios.

{
    \small
    \bibliographystyle{ieeenat_fullname}
    \bibliography{main}
}


\end{document}